# A Statistical Approach to Increase Classification Accuracy in Supervised Learning Algorithms

Valencia-Zapata; Gustavo A[1,3], Mejia; Daniel[1], Klimeck; Gerhard[1], Zentner; Michael G[1,2], and Ersoy; Okan[3].

**Abstract:** ***Probabilistic mixture models have been widely used for different machine learning and pattern recognition tasks such as clustering, dimensionality reduction, and classification. In this paper, we focus on trying to solve the most common challenges related to supervised learning algorithms by using mixture probability distribution functions. With this modeling strategy, we identify sub-labels and generate synthetic data in order to reach better classification accuracy. It means we focus on increasing the training data synthetically to increase the classification accuracy.***

**Index Terms:** ***Probabilistic mixture models, supervised learning, MNIST database, Bernoulli mixture model, EM algorithm, Maximum likelihood Estimator, Akaike information criterion, Bootstrapping.***

## 1. Introduction

THERE are numerous supervised learning algorithms, each of them with strengths and weaknesses. For instance, classical algorithms such as the Naive Bayes, linear classifiers or the widely used K-Nearest Neighbors (KNN) are among them. On the other hand, most recent methods such as Support Vector Machines, Neural Networks (NN), Convolutional Neural networks (CNN), etc have increasing popularity. Even though each algorithm represents a particular way to solve classification problems, there exist some common challenges related to supervised learning. Mainly, there can be cited five relevant challenges [1]: the bias-variance tradeoff, the amount of training data, class imbalance, data in high-dimensional space, and noisy labels. This work focuses on avoiding the effect of the bias-variance tradeoff and the amount of training data, identifying new labels (sub-labels based on the originals), and generating synthetic data for the training stage in order to improve the classification accuracy.

We show how our claims work with the MNIST database [2], which is widely used in the machine learning field and contains handwritten digits. It has a training set of 60000 digits and a test set of 10000 digits related to 10 classes representing digits between 0 and 9. Every digit is represented by a vector of length 784 or by a matrix of size 28×28, corresponding to an image. Additionally, we implement the KNN and Multilayer Neural Network (MLN) algorithms in the statistical software R environment, making use of the packages Class [3] and Mxnet [4], respectively.

---

Valencia-Zapata Gustavo A is at School of Electrical and Computer Engineering, Purdue University (e-mail: gvalenc@purdue.edu).

[1]Network for Computational Nanotechnology Cyber Node, Purdue University.

[2]HUBzero Research Group, Information Technology at Purdue

[3]Electrical and Computer Engineering, Purdue University, and Bogazici University, Istanbul, Turkey as adjunct professor.

## 2. Challenges in Supervised Learning

### 2.1 The bias–variance tradeoff

Algorithms with high bias produce simpler models and avoid overfit. However, they cannot capture relevant details in the feature space. In contrast, algorithms with high variance produce complex functions that can represent very well the training dataset, and captures those details, but then this sensitivity can produce overfit. In addition, a high variance is a direct effect of data in high dimensional space. Therefore, a large number of training samples are necessary in order to reduce the variance and misclassification [5].

### 2.2 The amount of training data

Figure 1 shows the mean of the error for different sizes of samples. It is evident that the larger the sample size, the smaller the mean of error for both models. Nevertheless, limitations related to the amount of the training data are common in classification tasks, not to mention the problems about high dimensionality.

Ensemble modeling has recently been one of the most successful strategies in order to obtain lower error rates for classification, and smaller errors for prediction. For example, some techniques are based on resampling such as Bagging [6], [7] (Boostrap aggregating), Boosting [8], and Stacking [9]. In Bagging, each model in the ensemble is independent with an equal vote, and uses random subsets of the original training dataset. Boosting is a sequential learning algorithm in which the first algorithm uses the complete training dataset, while the following algorithms focus on the misclassified cases of the previous model. Finally, in stacking, each model is independent, and uses the complete training set for learning. Then, the outputs of each model are combined in a new dataset, which is used by a final algorithm.

The next two sections present a novel methodology based on selecting the best models for KNN and MLN (The bias-variance tradeoff), and finding a probabilistic model to identify sub-labels (reduce the variance). This additional information based on sub-labels helps increase the training set with synthetic data. Our approach claims that additional information identified via probabilistic modeling can be useful for classification tasks not only for a specific algorithm but also for an ensemble model.

## 3. Selecting the Best Model

In order to consider the limitation of the amount of training data, we selected 10000 digits as a training dataset from the original MNIST training set. The best model for the KNN and the MLN was defined using a training set of 8000 digits, and a validation set of 2000 digits.

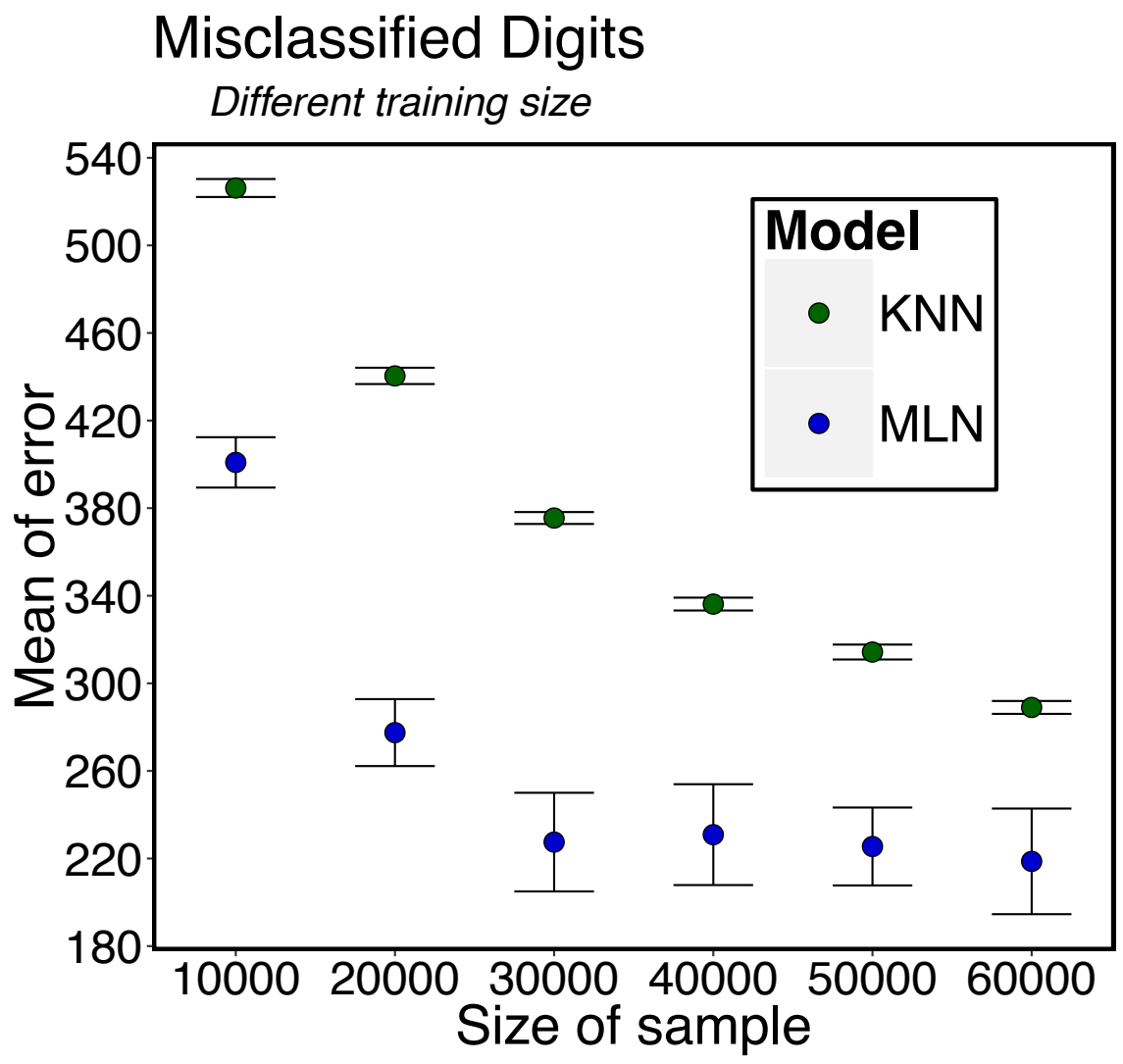

**Figure 1.** Mean of error for different training size for MNIST database.

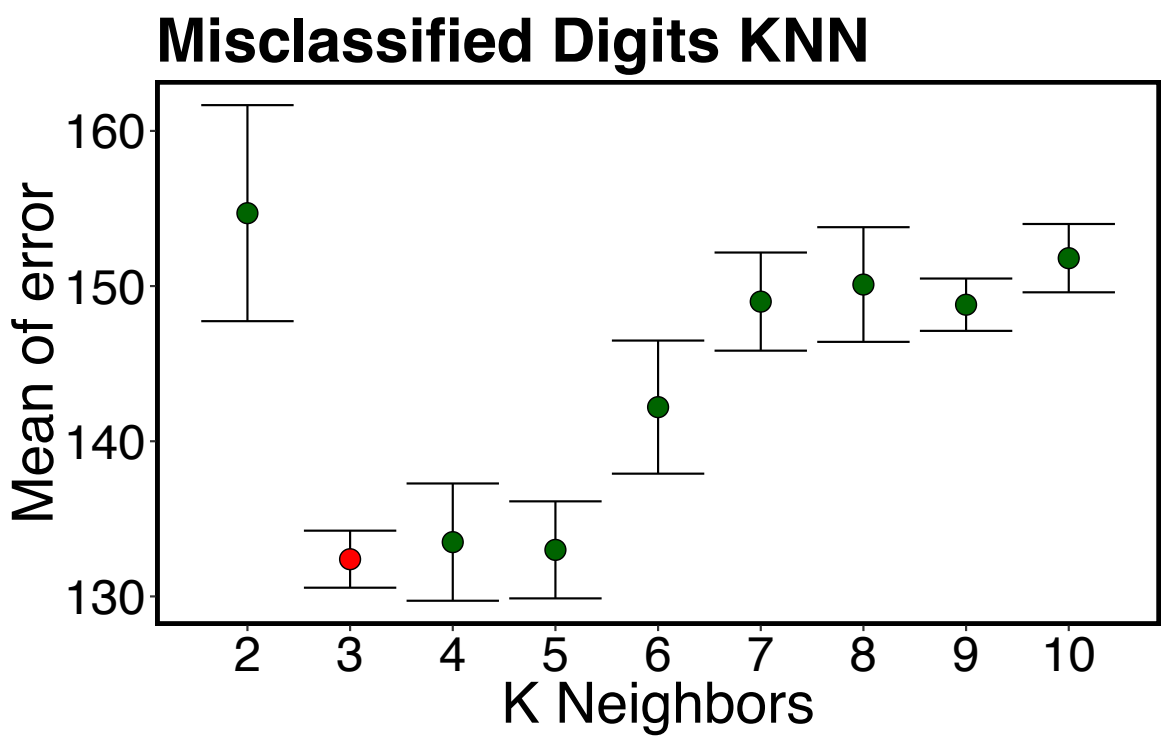

**Figure 2**. Mean of error for different number of neighbors.

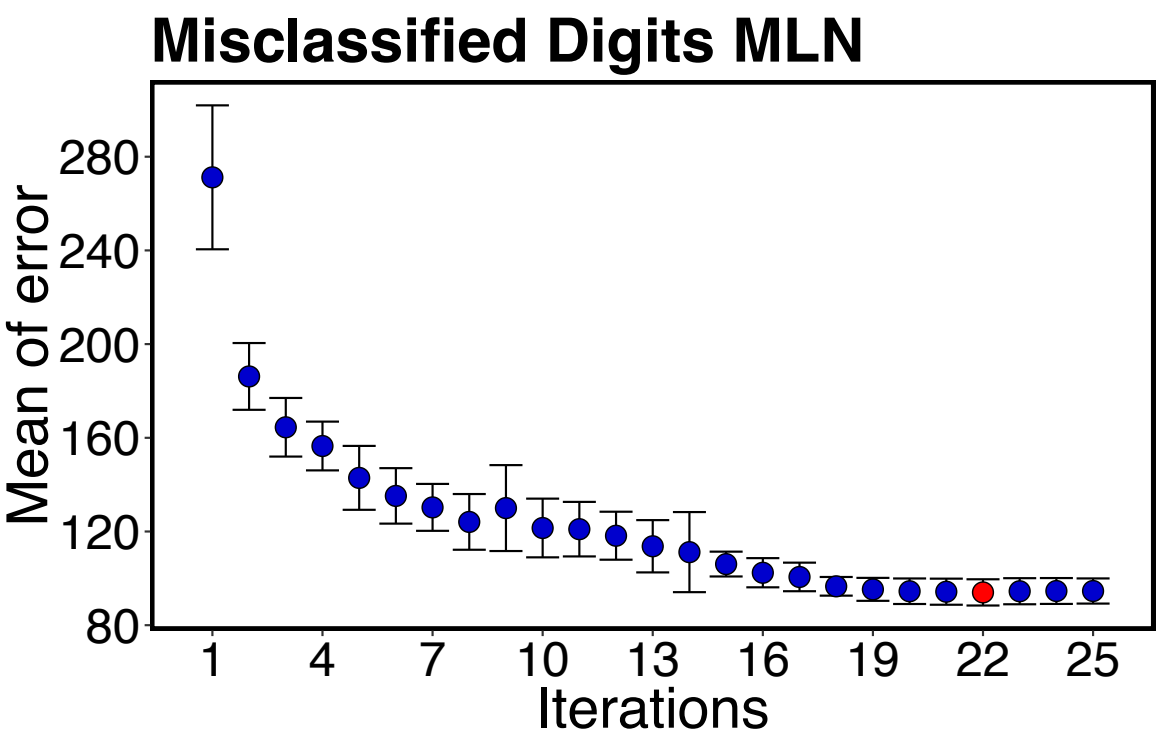

**Figure 3**. Mean of error Digits for different iterations.

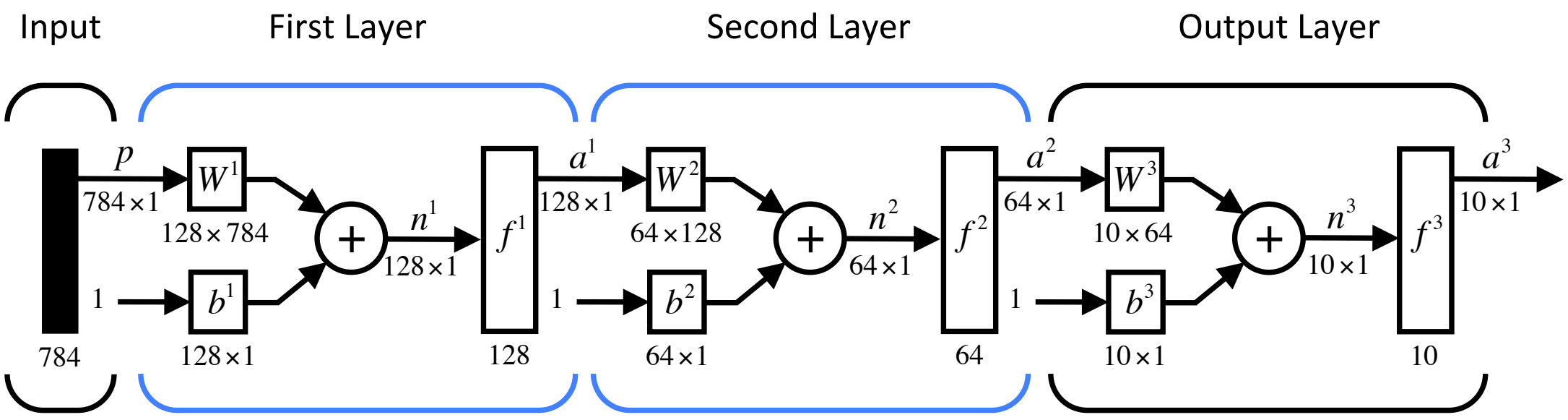

$$a^1 = f^1\left(W^1 p + b^1\right) \qquad a^2 = f^2\left(W^2 a^1 + b^2\right) \qquad a^3 = f^3\left(W^3 a^2 + b^3\right)$$

$$a^3 = f^3\left(W^3 f^2\left(W^2 f^1\left(W^1 p + b^1\right) + b^2\right) + b^3\right)$$

**Figure 4.** Multi-Layer Neural Network (MLN), Abbreviated Notation.

### *3.1 K-Nearest Neighbor*

KNN classifies the testing set based on the $K$ nearest training set vectors using the Euclidean distance [10]. Then, labels are decided by majority vote, and ties are broken randomly.

Choosing the value of $K$ is fundamental in order to reduce misclassification. A large $K$ value reduces the variance due to the noisy data; however, it produces a bias ignoring smaller patterns that can be useful for correct classification.

We selected $K = 3$, which is the number of neighbors with smallest mean error rate using $10$ different seeds. (Refer to Figure 2).

### *3.2 Neural Networks*

We trained a Multi-Layer Neural Network (MLN) using the well-known backpropagation algorithm [11], where the output is a probability vector with length equal to the number of labels. MLN performs nonlinear transformations in its hidden layers minimizing the squared error, which is reached using Gradient Descent [12].

The network is composed of the input layer, two hidden layers, and the output layer. Hidden layers have $128$ and $164$ neurons, respectively, and the output layer has 10 neurons associated with the same number of labels. The input $\mathrm{p}$ is represented by a vector of dimension $784{\times}1$, and each stage has its own weight matrix $\mathrm{W}$, bias vector $\mathrm{b}$, transfer function $\mathrm{f}$, and output vector $\mathrm{a}$. For example, the first layer has $128$ neurons. Thus, $\mathrm{W}^1$ is a $128 \times 784$ matrix, $\mathrm{b}^1$ is a vector of length $128$, and $\mathrm{a}^1$ is a vector of length $128$, which is the input for the second layer.

Rectified linear unit (ReLu) functions $\mathrm{f}^1$ and $\mathrm{f}^2$ are defined for smoothing the results $\mathrm{n}^1$ and $\mathrm{n}^2$, A Softmax function $\mathrm{f}^3$ is defined to obtain a probabilistic prediction vector $\mathrm{a}^3$, which allows classifying the input $\mathrm{p}$. Figure 4 shows the schematic for this MLN. The best model corresponds to the iteration $22$, which presents the smallest mean error rate using $10$ different seeds (Refer to Figure 3). The learning rate and momentum were set to 0.07 and 0.9, respectively.

### *3.3 The bias- variance for the study case*

Figure 5 shows the bias-variance for both algorithms and the testing dataset. The KNN algorithm presents high bias because several misclassified digits are recurrent for different seeds. That means seeds are mainly responsible for breaking ties. On the other hand, the MLN algorithm has high variance because several misclassified digits are not recurrent for all seeds. For instance, there are $40$ misclassified digits that only appear in one of the ten seeds. For the MLN algorithm, seeds determine the initialization of the weight matrices.

## 4. Probabilistic Mixture Model

The goal is to find the parameters of a mixture distribution that correspond the probability distribution of the training data set. The most used mixture models are based on Gaussian, Bernoulli, and Weibull distributions. Their applications extend to several fields such as market segmentation, binary images, text classification, DNA data, and reliability.

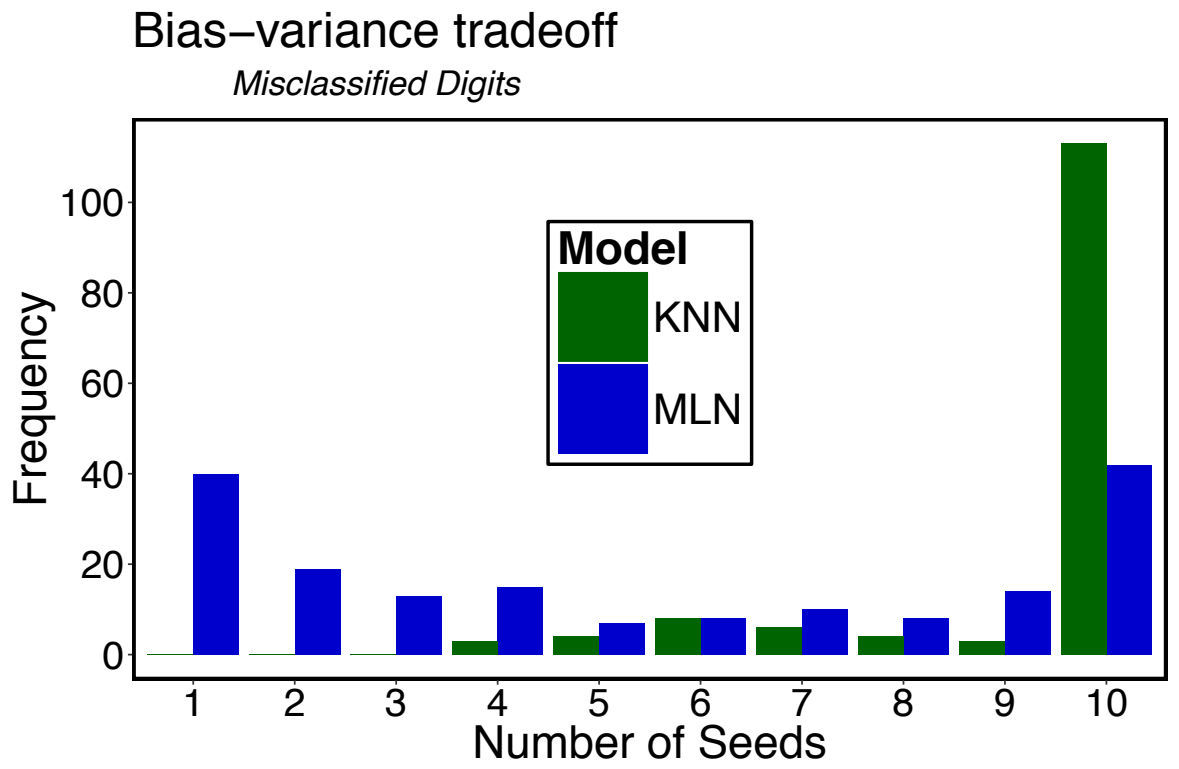

**Figure 5.** Bias-variance tradeoff for KNN and MLN

This work uses a Mixture of Bernoulli distributions. For this purpose, a threshold is used in order to binarize the training set, which originally is defined in grayscale $(0-255)$. For example, If the value of a pixel is equal to or greater than $100$, then this pixel becomes $1$ otherwise $0$.

Consider a set of $\mathrm{D}$ binary variables $\mathrm{x}_\mathrm{i}$, where $\mathrm{i} = (1,2, \dots \mathrm{D})$, each of which is governed by a Bernoulli distribution with parameter $\mathrm{p}_\mathrm{i}$. The set $\mathrm{D}$ represents the sequence of pixels of each digit, so that,

$$\mathrm{P}(\mathbf{x}|p) = \prod_{i=1}^{D} p_i^{x_i} (1-p_i)^{(1-x_i)} \qquad (1)$$

where $\mathbf{x} = (x_1, \dots, x_D)^T$ and $p = (p_1, \dots, p_D)^T$. We assume $x_i$'s are independent, given $p$. Therefore, $(2)$ is a finite mixture of these distributions:

$$\mathrm{P}(\mathbf{x}|p, \pi) = \sum_{k=1}^{K} \pi_k \mathrm{P}(\mathrm{x}|p_k) \qquad (2)$$

where $p = \{p_1, \dots, p_K\}$ and $\pi = \{\pi_1, \dots, \pi_K\}$ is the mixing coefficient vector. Using (1), we obtain

$$\mathrm{P}(\mathbf{x}|p_k) = \prod_{i=1}^{D} p_{ki}^{x_i} (1-p_{ki})^{(1-x_i)} \qquad (3)$$

Finally, for a given training data set $\boldsymbol{X} = \{\mathbf{x}_1, \dots, \mathbf{x}_\mathrm{N}\}$, the data log-likelihood function of the finite mixture, and the EM algorithm are used to estimate the mixture parameters [13]. Note that $\boldsymbol{X}$ is the set of digits for the training stage, with $N = 10000$, and $\hat{\mathcal{L}}$ is the Maximum Likelihood Estimator (LME) Equation(4). Its log-likelihood function is expressed by Equation (5), which estimates the probability of membership of a digit to each of $K$ labels.

$$\hat{\mathcal{L}}(\mathbf{X};\theta) = \mathrm{P}(\mathbf{X}|p,\pi) \quad (4)$$

$$\log \mathrm{P}(\mathbf{X}|p,\pi) = \sum_{n=1}^{N} \log \sum_{k=1}^{K} \pi_k \mathrm{P}(\mathbf{x}_n|p_k) \quad (5)$$

Even though the MNIST database has 10 labels, the goal is to reduce the variance through discovering sub-labels. For instance, several sub-labels may represent the same digit. Increasing the number of labels $K$ in the mixture model results in an increase in the dimensionality of the model, causing a monotonous increase in its likelihood.

The best mixture model is the one that maximizes the Akaike Information Criterion (AIC) [14], which seeks to balance the increase in likelihood and the complexity of the model by introducing a penalty term for each parameter. AIC is defined by

$$\mathrm{AIC}(\mathbf{X};\theta) = 2\eta(\theta) - 2\log \hat{\mathcal{L}}(\mathbf{X};\theta) \quad (6)$$

where $\eta(\theta)$ is the number of free parameters in the model that represents the complexity of the model. Then, for a Bernoulli Mixture Model (BMM) the $\eta(\theta)_{BMM}$ is defined by [15]

$$\eta(\theta) = K(D+1) - 1 \quad (7)$$

We initialized the EM algorithm [13] using $10$ different seeds and increased the number of labels $K$. Figure 5 shows this result for one of the seeds. For each number of classes ($K$ labels) plotted on the $X$ axis, the values of the MLE and AIC are plotted on the $Y$ axis. Finally, the best model that maximizes the AIC is reached when $K = 145$ and its value of AIC is $-1379830.6$. Therefore, given the training data set of $10000$ digits, the best number of labels is $145$.

## 5. BOOTSTRAPPING FOR STRONG SUB-LABELS

The "Purity" concept measures the degree of exclusivity inside sub-labels. A sub-label has Purity equal to 1 when its digits have the same original label. For instance, Figure 6 (a) shows the Centroid 30 (visual representation of sub-label30), which is composed of two types of original labels and its Purity is equal to 0.97. Similarly, Figure 6 (b) presents the Centroid 132, which is composed of seven original labels and its Purity is equal to 0.87.

Purity is estimated, and the label with the largest frequency is identified to represent each sub-label. Thus, sub-labels 30 and 132 are identified with label 8.

Several modeling strategies can be implemented using information related to strong sub-labels. For instance, resampling techniques can be used to produce synthetic data based on a mixture of a probabilistic distribution. Similar to the boosting technique, this model strategy links the misclassified digits detected in the validation set with the 145 sub-labels identified via mixture of distribution.

Table 1 shows the percentage of misclassified digits related to the validation data from Section 3. Even though the error rate for MLN is better than KNN, both algorithms present similar percentage of error for original labels. For instance, labels 8 and 3 present the larger number of misclassifications.

Table 2 shows seven strong sub-labels related to the original label 8. Using bootstrapping [16], we conducted resampling with replacement over the seven sub-labels 100 times.

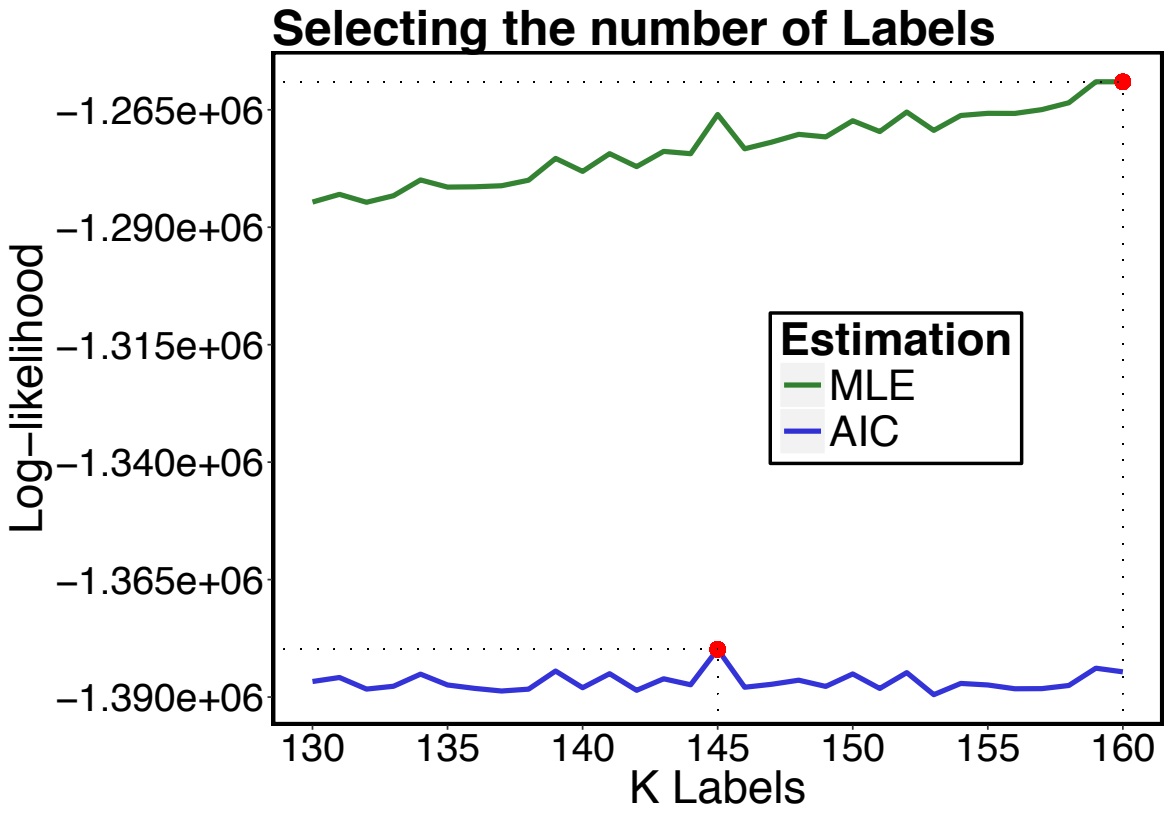

**Figure 5.** Selection of the best model using AIC.

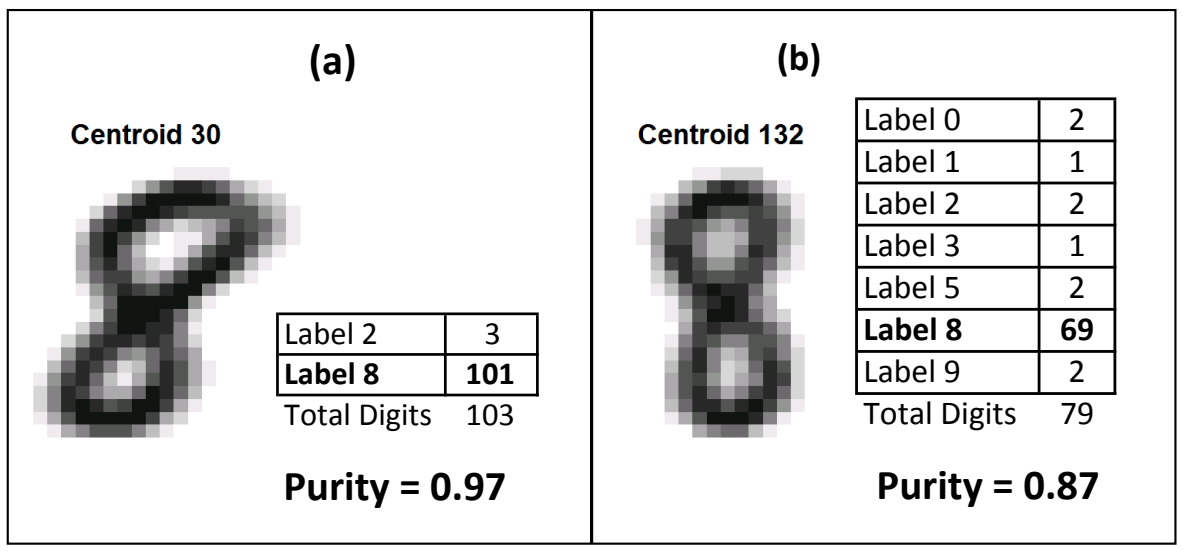

**Figure 6. An** example of Purity for two sub-labels.

| Percentages of Misclassified digits | | |
|---|---|---|
| **Label** | **KNN** | **MLN** |
| 0 | 0.03 | 0.06 |
| 1 | 0.01 | 0.03 |
| 2 | 0.12 | 0.11 |
| 3 | **0.17** | **0.17** |
| 4 | 0.10 | 0.08 |
| 5 | 0.09 | 0.11 |
| 6 | 0.02 | 0.06 |
| 7 | 0.06 | 0.06 |
| 8 | **0.27** | **0.17** |
| 9 | 0.08 | 0.10 |

**Table 1.** Percentage of error for each digit KNN and MLN algorithms.

In each resampling, two digits are selected randomly from the label with the largest frequency. Then, the mean is calculated producing the new synthetic digit. Finally, we produced 700 synthetic digits (100 digits by sub-label) related to the original label 8. Figures 7 and 8 compare the effect of these new digits over classification results for KNN and MLN. The following are the features of each case:

- $A =$ 10000 real digits (from MNIST) composing the training dataset.
- $B =$ 10000 real digits (from MNIST) + 700 synthetic digits (from sub-labels) composing the training dataset.
- $C =$ 10700 real digits (from MNIST) composing the training dataset

Training datasets in cases $A$ and $C$ are composed of real digits from the MNIST database. Case $B$ reduces the misclassified digits in case $A$ adding 700 synthetic digits related to label 8.

| Booststrapping Sub-labels Label 8 | | |
|---|---|---|
| **Sub-label** | **Num. of Digits 8** | **Purity** |
| 30 | 101 | 0.97 |
| 6 | 40 | 0.93 |
| 100 | 98 | 1 |
| 105 | 67 | 0.97 |
| 126 | 66 | 0.91 |
| 132 | 69 | 0.87 |
| 136 | 83 | 0.96 |

**Table 2.** Strong sub-label for label 8 .

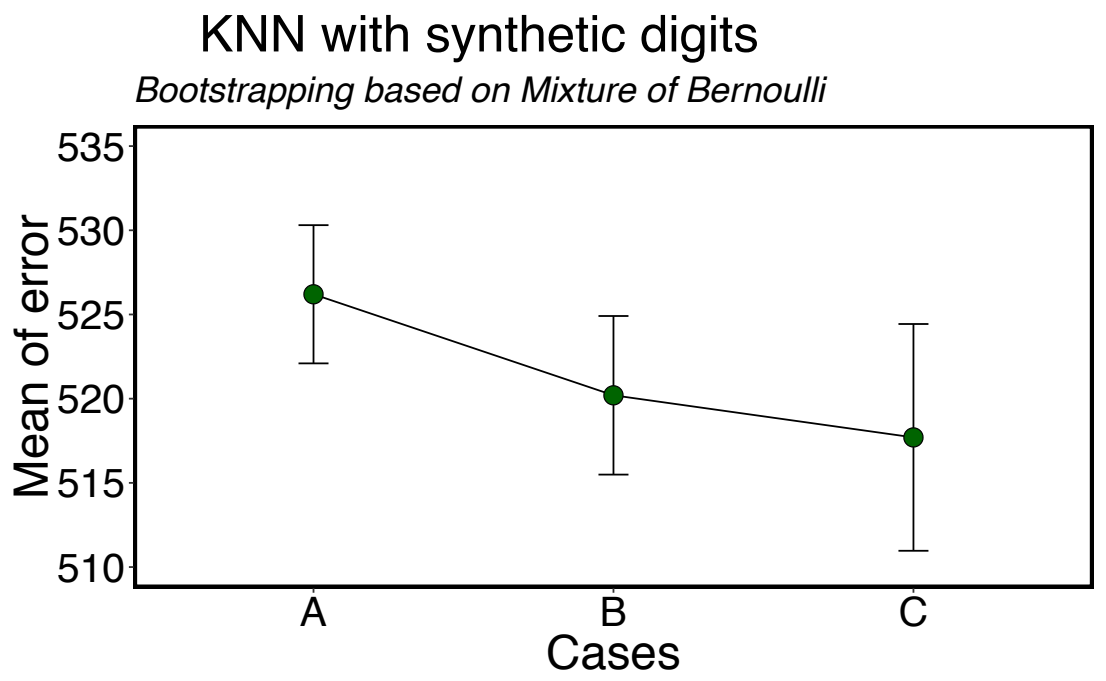

**Figure 7.** KNN comparison between real and real + synthetic data using Bootstrapping and Strong labels.

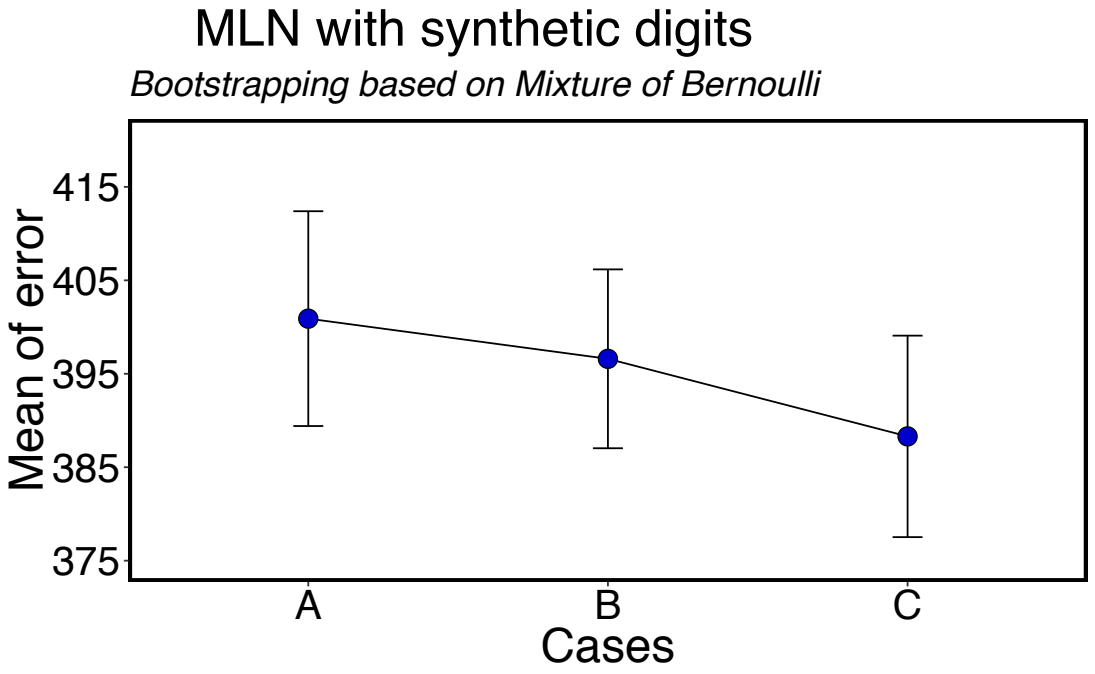

**Figure 8.** MLN comparison between real and real + synthetic data using Bootstrapping and Strong labels.

## 6. CONCLUSIONS

We proposed mixture of probabilistic distributions as a valid alternative for discovering new information related to labels in a classification problem. This new information allows reducing the variability by generating synthetic data through resampling with probability distributions.

Our statistical approach for increasing the amount of training data provides an additional understanding of each label, and shows good results not only for a high bias model such as KNN but also for a high variance model such as MLN. In both algorithms, the best model was identified using a validation dataset. Then, these models were trained with real data from MNIST database with additional synthetic data. The results showed improvement of classification accuracy with the addition of synthetic data in both algorithms.

One interesting conclusion of this work is the fact that increasing the training dataset based on probabilistic distributions reduces the number of misclassifications, even if the classification algorithm's parameters determining its architecture are unchanged. Hence, the focus of this research is independent of issues about the best architecture for the classification algorithm or tuning parameters to solve a particular problem.

Some directions for future work are to assess the effect of the proportion of real and synthetic data in the training dataset, class imbalance which may be produced by synthetic data, and increasing the number of classes. The latter may require modification of the classification algorithm. For instance, increasing the number or neurons or layers in MLN.

**Gustavo A Valencia-Zapata** (B.S. Electronic Engineering'04–G.C in Management'10–G.C in Statistics'11–MSc in Statistics'15) is a PhD student at the School of Electrical and Computer Engineering, Purdue University. He is as research assistant at the Network for Computational Nanotechnology Cyber Node, Purdue University. His research is mainly focused on developing machine learning algorithms, statistical, and data mining models in order to discover meaningful patterns in complex engineering problems.